\documentclass{article}

\PassOptionsToPackage{numbers}{natbib}

    \usepackage[preprint]{neurips_2019}

\usepackage[utf8]{inputenc} 
\usepackage[T1]{fontenc}    
\usepackage{hyperref}       
\usepackage{url}            
\usepackage{booktabs}       
\usepackage{amsfonts}       
\usepackage{nicefrac}       
\usepackage{microtype}      

\usepackage{dirtree}
\usepackage{graphicx}
\usepackage[multiple]{footmisc}

\usepackage{listings}
\usepackage{xcolor}
 
\definecolor{codegreen}{rgb}{0,0.6,0}
\definecolor{codegray}{rgb}{0.5,0.5,0.5}
\definecolor{codepurple}{rgb}{0.58,0,0.82}
\definecolor{backcolour}{rgb}{0.95,0.95,0.92}
 
\lstdefinestyle{mystyle}{
    backgroundcolor=\color{backcolour},   
    commentstyle=\color{codegreen},
    keywordstyle=\color{magenta},
    numberstyle=\tiny\color{codegray},
    stringstyle=\color{codepurple},
    basicstyle=\ttfamily\footnotesize,
    breakatwhitespace=false,         
    breaklines=true,                 
    captionpos=b,                    
    keepspaces=true,                 
    numbersep=5pt,                  
    showspaces=false,                
    showstringspaces=false,
    showtabs=false,                  
    tabsize=2
}
 
\title{rlpyt: A Research Code Base for Deep Reinforcement Learning in PyTorch}

\author{%
  Adam Stooke \\
  University of California, Berkeley \\
  \texttt{adam.stooke@berkeley.edu} \\
   \And
   Pieter Abbeel \\
   University of California, Berkeley \\
   \texttt{pabbeel@cs.berkeley.edu} \\
}

\begin{document}

\maketitle

\begin{abstract}
    Since the recent advent of deep reinforcement learning for game play \cite{DQN} and simulated robotic control (\textit{e.g.} \cite{trpo}), a multitude of new algorithms have flourished.  Most are model-free algorithms which can be categorized into three families: deep Q-learning, policy gradients, and Q-value policy gradients.  These have developed along separate lines of research, such that few, if any, code bases incorporate all three kinds.  Yet these algorithms share a great depth of common deep reinforcement learning machinery.  We are pleased to share \textit{rlpyt}, which implements all three algorithm families on top of a shared, optimized infrastructure, in a single repository.   It contains modular implementations of many common deep RL algorithms in Python using PyTorch \cite{pytorch}, a leading deep learning library.  rlpyt is designed as a high-throughput code base for small- to medium-scale research in deep RL.  This white paper summarizes its features, algorithms implemented, and relation to prior work, and concludes with detailed implementation and usage notes.  rlpyt is available at \texttt{https://github.com/astooke/rlpyt}. 
\end{abstract}

\section{Introduction}
Since the advent of deep reinforcement learning for game play in 2013 \cite{DQN} and simulated robotic control shortly after (\textit{e.g.} \cite{trpo}), a multitude of new algorithms have flourished.  Most are model-free algorithms which can be categorized into three families: deep Q-learning, policy gradients, and Q-value policy gradients.  These have developed along separate lines of research, such that few, if any, code bases incorporate all three kinds.  In fact, many of the original implementations remain unreleased.  As a result, practitioners often must develop from different starting points and potentially learn a new code base for each algorithm of interest or baseline comparison.  RL researchers often reimplement algorithms--perhaps a valuable individual exercise, but one that incurs redundant effort across the community, or worse, one that presents a barrier to entry.  Yet these algorithms share a great depth of common deep reinforcement learning machinery.  We are pleased to share \textit{rlpyt}, which implements all three algorithm families built on a shared, optimized infrastructure, in a single repository.   rlpyt contains modular implementations of many common deep RL algorithms in Python using PyTorch \cite{pytorch}, a leading deep learning library. Among numerous existing implementations, rlpyt
is a more comprehensive open-source resource for researchers.  rlpyt is available at \texttt{https://github.com/astooke/rlpyt}.

rlpyt is designed as a high-throughput code base for small- to medium-scale research in deep RL (large-scale being DeepMind AlphaStar \cite{alphastar} or OpenAI Five \cite{OpenAI_dota}, for example).  This white paper summarizes its features, algorithms implemented, and relation to prior work.  A small selection of learning curves are provided to verify learning performance for some standard RL environments in discrete and continuous control.  Notably, rlpyt reproduces record-setting results in the Atari domain from “Recurrent Experience Replay in Distributed Reinforcement Learning” (R2D2) \cite{r2d2}.  This benchmark requires on the order of 30 billion frames of game play and 1 million network updates, which rlpyt achieves in reasonable time without the use of distributed compute infrastructure.
Compatibility with the OpenAI Gym interface provides access to many existing learning environments and allows new ones to be freely customized.
This paper also introduces the "namedarraytuple", a new data structure for handling collections of arrays, which may be of outside interest.  Finally, more detailed implementation and usage notes are provided.

\subsection{Key Features and Algorithms}
Key capabilities and features include:
\begin{itemize}
    \item Run experiments in serial mode (helpful for debugging, sufficient for some experiments).
    \item Run experiments parallelized, with options for parallel sampling and/or multi-GPU optimization.
    \item Sampling and optimization synchronous or asynchronous (via replay buffer).
    \item Use CPU or GPU for training and/or batched action selection during environment sampling.
    \item Full support for recurrent agents.
    \item Online or offline evaluation and logging of agent diagnostics during training.
    \item Includes launching utilities for stacking / queueing sets of experiments on local computer.
    \item Modularity for easy modification and re-use of existing components.
    \item Compatible with OpenAI Gym \cite{openai_gym} environment interface.\footnote{See implementation details for required modification.}
\end{itemize}

Implemented algorithms include the following (check the repository for possible additions):
\begin{itemize}
    \item \textbf{Policy Gradient}: A2C \cite{A3C}, PPO \cite{PPO}.
    \item \textbf{Deep Q-Learning}: DQN \cite{DQN} + variants: Double \cite{doubledqn}, Dueling \cite{dueling}, Categorical \cite{CatDQN}, Rainbow \cite{rainbow} (minus Noisy Nets), Recurrent (R2D2-like) \cite{r2d2}, including vector-valued epsilon-greedy (Ape-X-like) \cite{ape-x} (\textit{coming soon}: Implicit Quantile DQN \cite{iqn}).
    \item \textbf{Q-Function Policy Gradient}: DDPG \cite{ddpg}, TD3 \cite{td3}, SAC \cite{sac, SAC2}, (\textit{coming soon}: Distributional DDPG \cite{d4pg}).
\end{itemize}

Replay buffers support both the DQN and Q-function policy gradient algorithms and include the following options: n-step returns; sequence replay (for recurrence); periodic storage of recurrent state (to save memory); prioritized replay (sum tree) \cite{prioritized}; frame-based buffer, to save memory e.g. by storing only unique Atari frames.

\section{Parallel Computing Infrastructure for Faster Experimentation}

The two phases of model-free RL--sampling environment interactions and training the agent--can be parallelized differently.  rlpyt addresses both, as described here.  In all arrangements, system shared memory underlies inter-process communication of training data and model parameters, minimizing data transfer time and memory footprint.

\subsection{Sampling}
\begin{figure}[h]
  \centering
  \includegraphics[width=0.9\textwidth]{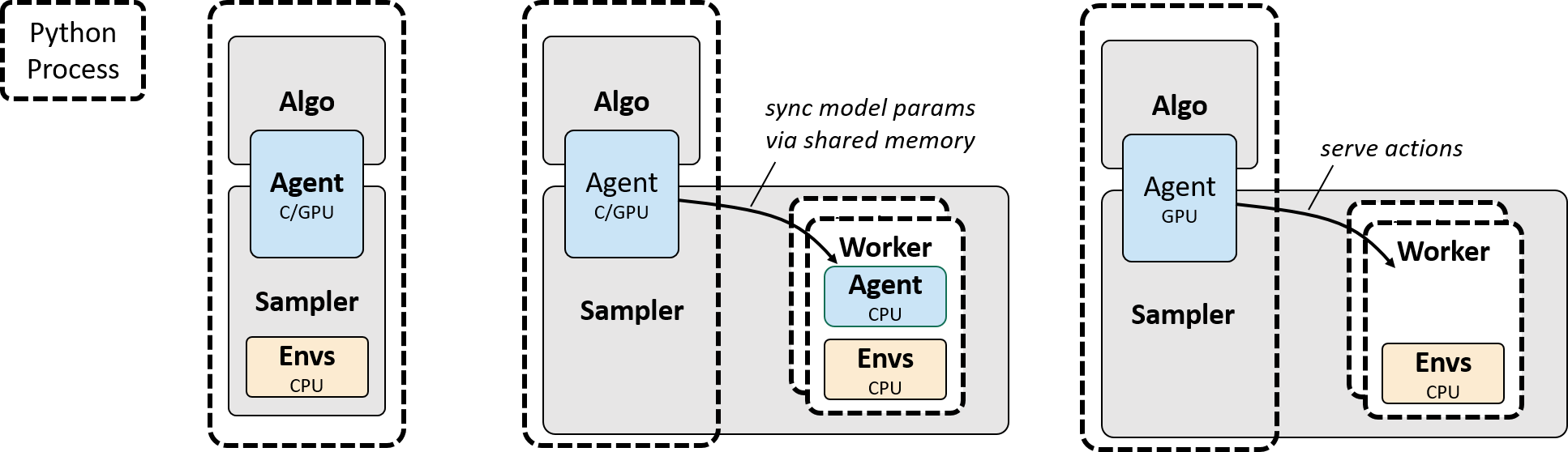}
  \caption{Environment interaction sampling schemes.  \textit{(left)} Serial: agent and environments execute within one Python process.  \textit{(center)} Parallel-CPU: agent and environments execute on CPU in parallel worker processes.  \textit{(right)} Parallel-GPU: environments execute on CPU in parallel workers processes, agent executes in central process, enabling batched action-selection.}
  \label{fig:sampling}
\end{figure}

For sampling, rlpyt offers the following configurations, also depicted in Figure \ref{fig:sampling}.

\textbf{Serial.}  Sampling occurs in the master process and can run one or more environment instances.  The built-in agent uses the same model for sampling and for optimization, so if optimizing on the GPU, action-selection during sampling also uses the GPU, batched over all environments.  (If running many time steps per batch with GPU optimization, it may be faster to use a separate CPU model for action-selection.)

\textbf{Parallel-CPU.}  The sampler launches worker processes to run environments and perform action selection.  If optimizing on the GPU, model parameters are copied to shared memory for CPU action selection in workers.  Synchronization across workers only occurs per sampling batch.

\textbf{Parallel-GPU.}  The sampler launches worker processes to run environments, and observations are communicated back to the master process for action selection, which will use the GPU if optimizing on GPU.  All the environments' observations are batched together for one call to the agent.  Step-wise communication happens via another shared memory buffer, and light-weight semaphores enforce synchronization across workers at every simulation batch-step.

\textbf{Alternating-GPU.}  Like parallel-GPU sampling but with two groups of workers; one group steps environments while the other group awaits action-selection.  May provide speedups when the action-selection time is similar to but shorter than the batch environment simulation time.

\vspace{2mm}

\subsection{Optimization}
\begin{figure}[h]
  \centering
  \includegraphics[width=0.75\textwidth]{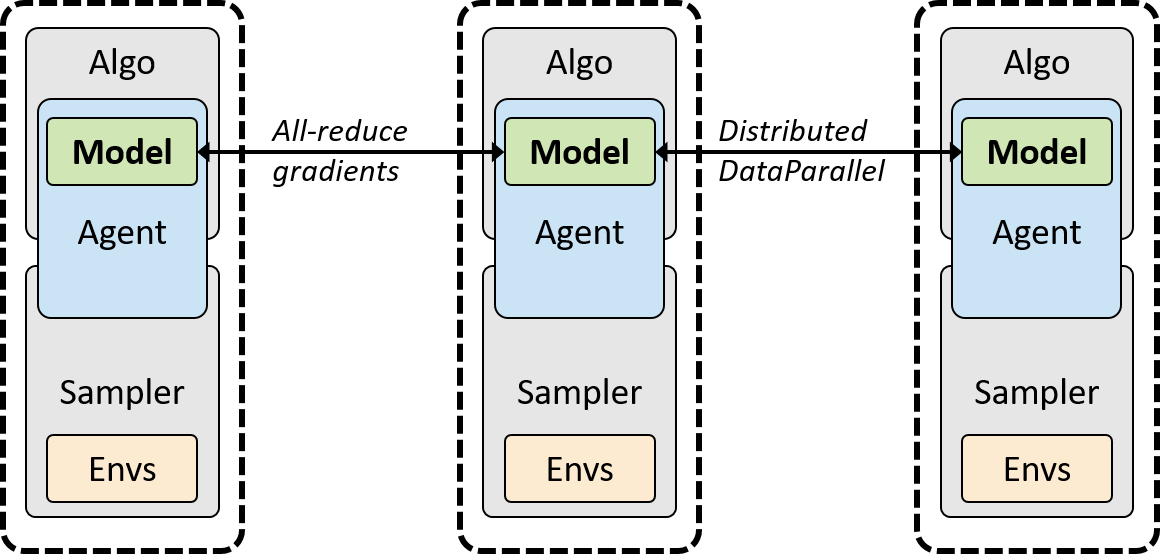}
  \caption{Synchronous multi-process reinforcement learning.  Each python process runs a copy of the full sampler-algorithm stack, with synchronization enforced implicitly during backpropagation in PyTorch’s \texttt{DistribuedDataParallel} class.  Both GPU (NCCL backend) and CPU (gloo backend) modes are supported.}
  \label{fig:sync}
\end{figure}

Synchronous multi-GPU optimization is implemented using PyTorch’s \texttt{DistributedDataParallel} to wrap the model.  A separate python process drives each GPU.  As provided by PyTorch, NCCL is used to all-reduce every gradient, which can occur in chunks concurrently with backpropagation, for better scaling on large models. The same applies for multi-CPU optimization, using \texttt{DistributedDataParallelCPU} and the “gloo” backend (may be faster than MKL threading for multiple CPU cores).  The arrangement is shown in Figure \ref{fig:sync}.  The entire sampling-training stack is replicated in each process and no training data is shared among them.  Any of the serial or parallel samplers can be used.

\vspace{2mm}

\subsection{Asynchronous Sampling-Optimization}
\begin{figure}[h]
  \centering
  \includegraphics[width=0.5\textwidth]{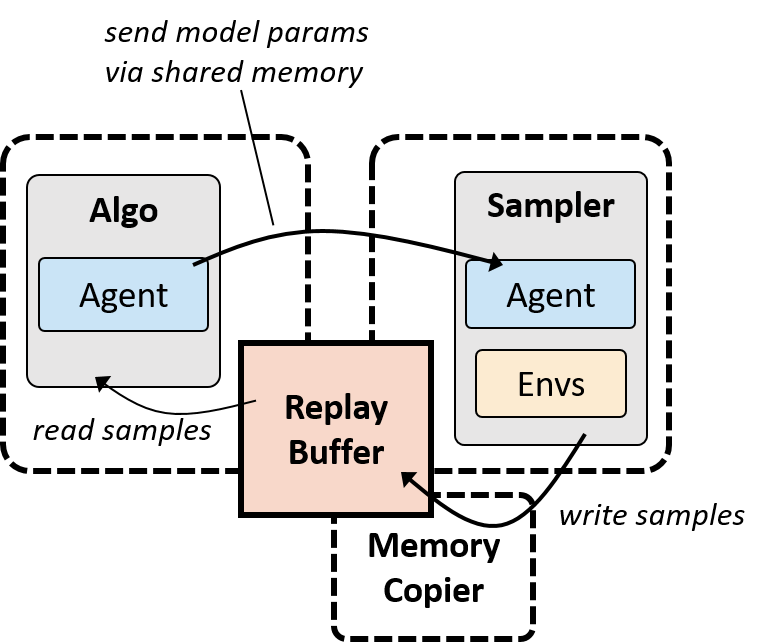}
  \caption{Asynchronous sampling/optimization mode.  Separate python processes run optimization and sampling via a shared-memory replay buffer under read-write lock.  Memory copier processes write from the sampler batch buffer (a double buffer) to the replay buffer, freeing the sampler to proceed immediately from batch to batch of collection.}
  \label{fig:async}
\end{figure}

In the configurations depicted so far, the sampler and optimizer operate sequentially in the same Python process.  In some cases, however, running optimization and sampling asynchronously achieves better hardware utilization, by allowing both to run continuously. In asynchronous mode, separate Python processes run the training and sampling, tied together by a replay buffer built on shared memory.  Sampling runs uninterrupted by the use of a double buffer for data batches, which yet another Python process copies into the main buffer, under a read-write lock.  This is shown in Figure \ref{fig:async}.  The optimizer and sampler may be parallelized independently, perhaps each using a different number of GPUs, to achieve best overall utilization and speed.

Some level of control between the processes is maintained.  A desired maximum replay ratio can be specified (rate of consumption divided by rate of generation of training data), and the optimizer will be throttled not to exceed this value.  The sampler batch size (time-steps) determines rate of actor model update, if new parameters are available from the optimizer.  All actors use the same parameters.

\subsection{Which Configuration is Best?}
When creating or modifying agents, models, algorithms, and environments, serial mode will be the easiest for debugging.  Once that runs smoothly, it is straightforward to explore the more sophisticated infrastructures for parallel sampling, multi-GPU optimization, and asynchronous sampling, since they are built on largely the same interfaces.  Of course, deviations from the standard RL work-flow (i.e. the \textit{runner}) may require more care to parallelize--again it is recommended to start with the serial case.  The optimal configuration may depend on the problem, available compute hardware, and the number of experiments to run.  Currently, rlpyt implements only single-node parallelism, but its components could form building blocks for a distributed framework.

\section{Learning Performance}
This section presents learning curves which verify the performance of the implementations against published values.  A subset of standard Atari games \cite{bellemare2013arcade} and Mujoco \cite{mujoco} continuous control environments are shown.  This is neither a comprehensive benchmark nor guide to scaling, but merely an exercise of each algorithm and infrastructure component.  For Atari scaling guidelines, see e.g. \cite{accel_rl}, for Mujoco, \cite{d4pg} is a likely starting point.

\subsection{Mujoco: Continuous Control from State}

Here we present reinforcement learning algorithms applied to continuous control from state on a selection of Mujoco\footnote{mujoco200.} tasks in OpenAI Gym.  For each algorithm, we used the same published hyperparameters across all environments and ran serial implementations.\footnote{A previous version of this paper showed lower scores for SAC and TD3, which have improved here by bootstrapping the value function when the trajectory ends due to time limit, as in the original SAC implementation.  Scores for SAC further improved by switching to the newer version, with entropy tuning and no state-value function.}

\begin{figure}[h]
  \centering
  \includegraphics[width=0.98\textwidth]{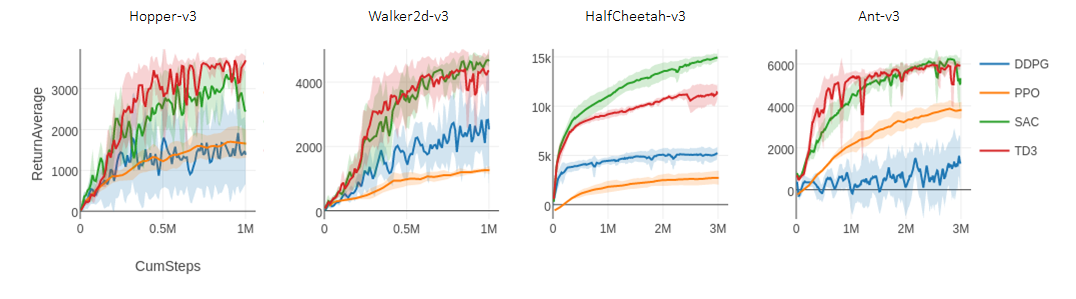}
  \caption{Continuous control in Mujoco by RL algorithms--DDPG (settings from \cite{td3}), TD3, SAC, and PPO; 4 random seeds each.} 
  \label{fig:mujoco}
\end{figure}

\subsection{Atari: Discrete Control from Vision}

\begin{figure}[h]
  \centering
  \includegraphics[width=0.98\textwidth]{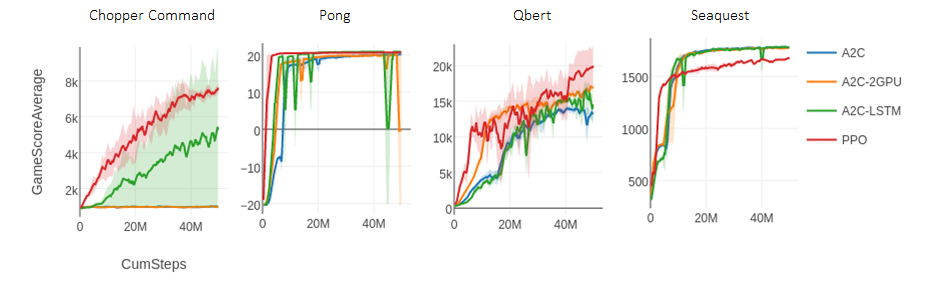}
  \caption{Policy gradient algorithms--A2C (feed-forward), A2C-LSTM (1-frame observation), A2C-2GPU (synchronous mode), PPO; 2 random seeds each.}
  \label{fig:pg_atari}
\end{figure}

\begin{figure}[h]
  \centering
  \includegraphics[width=0.98\textwidth]{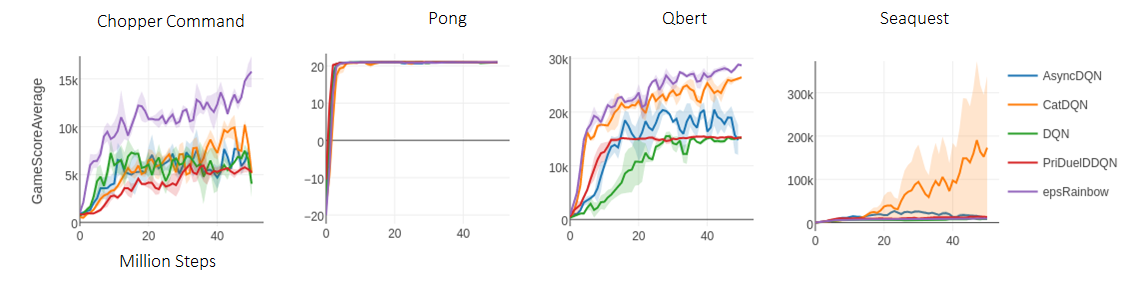}
  \caption{DQN plus variants--Categorical, Prioritized-Dueling-Double, Rainbow minus Noisy Nets, and asynchronous mode--all trained with batch size 128; 2 random seeds.}
  \label{fig:dqn}
\end{figure}

Here we include learning curves for a small selection of Atari games learned by vision using both policy gradient (Figure \ref{fig:pg_atari}) and DQN algorithms (Figure \ref{fig:dqn}).

\begin{figure}[h]
  \centering
  \includegraphics[width=0.98\textwidth]{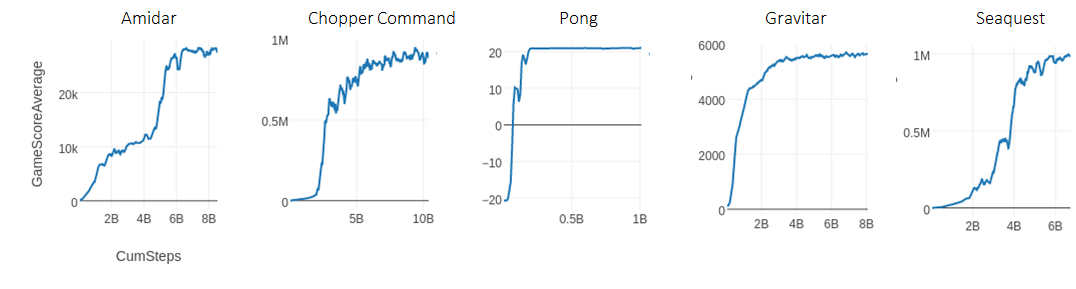}
  \caption{Reproduction of R2D2 learning curves in rlpyt, a single seed each.}
  \label{fig:r2d1}
\end{figure}

\textbf{R2D1} 
We highlight rlpyt's reproduction of the state of the art performance of R2D2 \cite{r2d2}, which was previously only feasible using distributed computing.  This benchmark includes a recurrent agent trained from a replay buffer for on the order of 10 billion samples (40 billion frames).  R2D1 (non-distributed R2D2) exercises several of rlpyt's more advanced infrastructure components to achieve this, namely multi-GPU asynchronous sampling mode with the alternating sampler.  In Figure \ref{fig:r2d1} we reproduce several learning curves which surpass any previous algorithm.  Some slight differences in performance against published values most likely resulted from a difference in the prioritization for new samples, which affected some games more than others,\footnote{Most curves used 1-step TD errors for prioritizing new samples and had unintentionally swapped the two replay priority coefficients.  Furthermore, since collection ran in 40 time-step batches but training used 80-step sequences, we used only half the training segment to compute new priorities.  \texttt{Gravitar} was especially sensitive and improved when we corrected to use 5-step TD initial priorities and by using the \textit{second} half-batch, yet this run still plateaued at a low score, below 6,000.  Work to remedy this continues.} and a slightly lower replay ratio.\footnote{We used a replay ratio of 1, including the warmup samples, whereas the original authors ran a replay ratio near 0.8 counting only the training samples; by their counting we ran at 0.67.}  Given the low replay ratio, initial priorities are very important in some games.

\begin{figure}[h]
  \centering
  \includegraphics[width=0.6\textwidth]{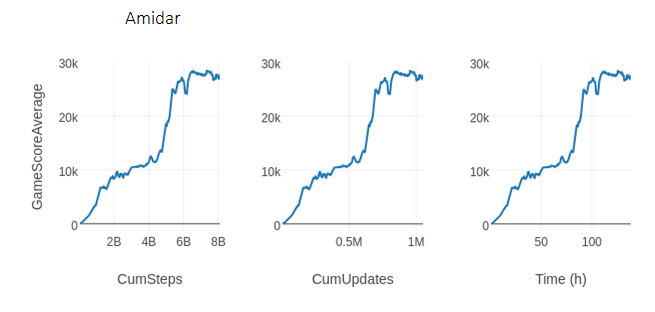}
  \caption{The same learning curve for the game \texttt{Amidar} over three horizontal axes: environment steps, model updates, and wall-clock time, for rlpyt's R2D1 implementation run in asynchronous sampling mode using 24 CPU cores and 3 GPUs.}
  \label{fig:amidar}
\end{figure}

The original, distributed implementation of R2D2 quoted about 66,000 steps per second (SPS) using 256 CPUs for sampling and 1 GPU for training.  rlpyt achieves over 16,000 SPS when using only 24 CPUs\footnote{2x Intel Xeon Gold 6126, circa 2017.} and 3 Titan-Xp GPUs in a single workstation (one GPU for training, two for action-serving in the alternating sampler).  This may be enough to enable experimentation without access to distributed infrastructure.  One possibility for future research is to increase the replay ratio (here set to 1) for faster learning using multi-GPU optimization.  Figure \ref{fig:amidar} shows the same learning curve over three different measures: environment steps (\textit{i.e.} 1 step = 4 frames), model updates, and time.  This run reached 8 billion steps and 1 million updates in less than 138 hours.  

\section{New Data Structure: \texttt{namedarraytuple}}
rlpyt introduces new object classes "namedarraytuples" for easier organization of collections of numpy arrays or torch tensors.  A namedarraytuple is essentially a namedtuple which exposes indexed or sliced read/writes into the structure.  Consider writing into a (possibly nested) dictionary of arrays which share some common dimensions for addressing:
 
\begin{minipage}{\textwidth}
\begin{lstlisting}[language=Python]
for k, v in src.items():
  if isinstance(dest[k], dict):
	  ..recurse..
  dest[k][slice_or_indexes] = v
\end{lstlisting}
\end{minipage}
This code is replaced by the following:
\begin{lstlisting}
dest[slice_or_indexes] = src
\end{lstlisting}

Importantly, the syntax is the same whether \texttt{dest} and \texttt{src} are individual numpy arrays or arbitrarily-structured collections of arrays.  The structures of \texttt{dest} and \texttt{src} must match, or \texttt{src} can be a single value to apply to all fields, and \texttt{None} is a special placeholder value for fields to ignore. rlpyt uses this data structure extensively--different elements of training data are organized with the same leading dimensions, making it easy to interact with desired time- or batch-dimensions.

This is also intended to support environments with multi-modal observations or actions. Rather than flattening and merging, say, camera images and joint-angles into one observation vector, the environment can store them as-is into a namedarraytuple for the observation.  In the forward method of the model, \texttt{observation.joint} and \texttt{observation.image} can be fed into the desired layers, without changing intermediate infrastructure code.  For more details, see the code and documentation for namedarraytuples in \texttt{rlpyt/utils/collections.py}.

The use of namedtuples and namedarraytuples may incur some programming overhead during setup or modification of agents, algorithms, and environments.  For example, for serialization\footnote{The only built-in use of serialization for samples data is the option of dropping into a subprocess while generating initial examples for buffer allocation.  Model forward execution triggers MKL OpenMP threading initialization which can affect subprocesses thereafter.  For example, parallel-CPU sampler agents should be initialized with 1 MKL thread if on 1 CPU core, whereas the optimizer might use multiple cores and threads.  Incidentally, most rlpyt subprocesses set \texttt{torch.num\_threads(1)} to avoid hanging on MKL, which might not be fork-safe.}  they must be defined at the module-level, which can be accomplished dynamically via the use of a global variable (see the Gym wrappers).  A benefit of these explicitly-defined interfaces is that they reduce chance of mistake by omission or replacement of a shared-memory buffer element by local memory.

\section{Related Work}
For newcomers to deep RL, other resources may be better for familiarization with algorithms, such as OpenAI Spinning Up \cite{spinningup}.\footnote{\texttt{https://spinningup.openai.com/en/latest/index.html}}\footnote{\texttt{https://github.com/openai/spinningup}} rlpyt is a revision and extension of the \textit{accel\_rl} codebase, \footnote{\texttt{https://github.com/astooke/accel\_rl}} which explored scaling RL in the Atari domain using Theano \cite{theano}, see \cite{accel_rl} for results.  For a further study of scaling in deep learning including RL, see \cite{mccandlish2018empirical}.  rlpyt and accel\_rl were originally inspired by \textit{rllab} \cite{rllab} (for example the logger remains nearly a direct copy)\footnote{\texttt{https://github.com/rll/rllab}}. 

Other published research code bases include OpenAI Baselines \cite{openaibaselines} and Dopamine \cite{dopamine}, both of which are implemented in Tensorflow \cite{tensorflow}, and neither of which are optimized to the extent of rlpyt nor contain all three algorithm families.  Rllib \cite{rllib}, built on top of Ray \cite{ray}, focuses on distributed computing, possibly complicating small experiments.  Facebook Horizon \cite{horizon} offers a subset of algorithms and focuses on applications toward production at scale.  In sum, rlpyt provides modular implementations of more algorithms and modular infrastructure for parallelism, making it a distinct tool set supporting a wide range of research uses.

\section{Implementation and Usage Details}
To get started, it is recommended to follow the example scripts provided in the repository and read the notes therein.  The following is a conceptual overview without code.

\subsection{Code Structure.}
The following tree and descriptions summarize the structure of classes and interfaces.

\begin{figure}[h!]
\centering
\begin{minipage}{6cm}
\dirtree{%
.1 Runner.
.2 Sampler.
.3 Collector.
.3 Environment.
.4 Observation Space.
.4 Action Space.
.3 TrajectoryInfo.
.2 Agent.
.3 Model.
.3 Distribution.
.2 Algorithm.
.3 Optimizer.
.3 OptimizationInfo.
.2 Logger.
}
\end{minipage}
\end{figure}

\textbf{Runner} - Connects the sampler, agent, and algorithm; manages the training loop and logging of diagnostics.

\textbf{Sampler} - Manages agent-environment interaction to collect training data; can initialize parallel workers.

\textbf{Collector} - Steps environments (and maybe operates agent) and records samples.

\textbf{Environment} - The task to be learned.  As in previous implementations, at each step outputs: \texttt{(observation, reward, done, env\_info)}.

\textbf{Observation/Action Space} - Interface specifications from environment to agent.

\textbf{TrajectoryInfo} - Diagnostics logged on a per-trajectory basis.

\textbf{Agent} - Chooses control action to the environment in sampler; trained by the algorithm; interface to model; holds model recurrent state during sampling.  As in previous implementations, at each step outputs \texttt{(action, agent\_info)}.

\textbf{Model} - PyTorch neural network module accepting \texttt{(observation, prev\_action, prev\_reward)} and possibly \texttt{initial\_rnn\_state} arguments.

\textbf{Distribution} - Samples actions for stochastic agents; defines related formulas for loss functions.

\textbf{Algorithm} - Uses gathered samples to train the agent, e.g. defines a loss function and performs gradient descent.

\textbf{Optimizer} - Training update rule (e.g. Adam) for model parameters.

\textbf{OptimizationInfo} - Diagnostics logged on a per-training batch basis.

\textbf{Logger} - Available throughout all processes and classes for recording printed statements and/or tabular values.

\subsection{No Asynchronous Optimization}  
Recent projects in large-scale RL, such as OpenAI Five \cite{OpenAI_dota} and DeepMind AlphaStar \cite{alphastar}, have succeeded using synchronous multi-device optimization (meaning every gradient is all-reduced across devices, which hold the same parameter values).  Previous experience in \cite{accel_rl} found good scaling of asynchronous, multi-GPU A3C and PPO on Atari using a CPU parameter store, but this technique did not scale as well to larger networks with more training updates, such as in DQN.  Therefore, rlpyt currently does not include asynchronous optimization schemes such as those in \cite{A3C, hogwild}.

\subsection{Recurrent Agents}
All agents receive the \texttt{(observation, previous\_action, previous\_reward)} inputs (see e.g. \cite{A3C}), although standard feedforward agents might use only the observation. The recurrent state is organized into its own namedarraytuple and can be customized.

\textbf{Sampling.}  The agent handles the recurrent state during environment sampling.  This functionality is provided in an optimized fashion according to the CuDNN \cite{cudnn} interface, agnostic to the structure of that state.  Separate mixin classes for custom agents are included for regular sampling and alternating sampling.  Recurrent state is recorded under \texttt{agent\_info}.

\textbf{Training.} Training data is organized with leading dimensions of \texttt{[Time, Batch]}, matching the PyTorch/CuDNN implementations of recurrence.  For CuDNN, the initial recurrent state must be re-organized into \texttt{[Num\_Layers, Batch, Hidden\_Size]} dimensions and made contiguous, as shown in the included recurrent agent classes.

\subsection{Data Organization Inferred in Model Forward Method}
The same model can be used with different leading dimensions: a single input (no leading dims), a batch \texttt{[Batch, ..]}, or a time-batch \texttt{[Time, Batch, ..]}.  In the model’s forward method, leading dimensions are inferred according to known dimension of the observation, for example.  Inputs are reshaped accordingly for feed-forward or recurrent layers, and finally the outputs have their leading dimensions restored according to what was input.  This way, the same model can be used for action-selection during sampling, for training, and for extracting single examples for constructing buffers.  See any of the included models for a template of this pattern which should be followed in any custom models.

\subsection{OpenAI Gym Interface}  
The use of preallocated buffers requires one modification to the Gym environment interface: the \texttt{env\_info} dictionary must provide the same keys/fields at every step.  A Gym-style wrapper is included, which converts the \texttt{env\_info} into a namedtuple for easy writing into the samples buffer.  An additional wrapper component is provided as one way to ensure all keys are present at every step.  A wrapper is also provided for Gym spaces to convert them to the corresponding rlpyt space (notably the multi-modal Gym \texttt{Dictionary} space becomes the rlpyt \texttt{Composite} space.)

\subsection{Launching Utilities}  Launching utilities are included for building variants and stacking / queueing experiments on given local hardware resources.  For example, on an 8-GPU, 40-CPU machine, one may want to run some number of variants (say, 30 different settings/seeds), each using 2 GPUs; the launcher will launch 4 experiments on non-overlapping resources (each with 2 GPUs and 10 CPUs), and as those finish, it will launch the next in their places until all are complete.  Results are recorded into a file structure which matches that of the variants generated (see the example scripts).  Other scripting patterns may be preferable for widely parallelized launching into the cloud.

\section{Conclusion}
We hope that rlpyt can facilitate adoption of existing deep RL techniques and serve as a launching point for research into new ones.  For example, the more advanced topics of meta-learning, model-based, and multi-agent RL are not explicitly addressed in rlpyt, but applicable code components may still be helpful in accelerating their development.  We expect the offering of algorithms to grow over time as the field matures.

\subsubsection*{Acknowledgments}
First we acknowledge the original authors of all algorithms and supporting libraries, as listed in the references.  Thanks to Steven Kapturowski for clarification of several implementation details of R2D2, and to Josh Achiam and Wilson Yan for help debugging SAC.  Adam Stooke gratefully acknowledges the support of the Fannie and John Hertz Foundation and the NVIDIA Corporation.

\bibliographystyle{unsrtnat}
\bibliography{main.bbl}

\end{document}